%% file: DetectabilityByGSD.tex
\documentclass[review,authoryear]{elsarticle}
\usepackage{natbib}
\usepackage{subfigure}
\usepackage{lineno}
\usepackage[colorlinks,urlcolor=blue]{hyperref}
\usepackage{amsmath,amssymb}
\usepackage{diagbox}
\usepackage{soul}
\usepackage{multirow}
\usepackage{graphicx} 
\usepackage{caption}
\usepackage{setspace}

\begin{document}
\begin{frontmatter}
	\title{Automated Aerial Animal Detection When Spatial Resolution Conditions Are Varied}
	\author{Jasper Brown$^{a*}$, Yongliang Qiao$^{a}$, Cameron Clark$^{b}$, Sabrina Lomax$^{b}$, Khalid Rafique$^{a}$, and Salah Sukkarieh$^{a}$}
	\address{$^{a}$Australian Centre for Field Robotics (ACFR), Faculty of Engineering, The University of Sydney, NSW 2006, Australia}
	\address{$^{b}$Livestock Production and Welfare Group, School of Life and Environmental Sciences, Faculty of Science, The University of Sydney, NSW 2006, Australia}
	\cortext[mycorrespondingauthor]{Corresponding author: j.brown@acfr.usyd.edu.au}
	
	\begin{abstract}
		Knowing where livestock are located enables optimized management and mustering. However, Australian farms are large meaning that many of Australia’s livestock are unmonitored which impacts farm profit, animal welfare and the environment. Effective animal localisation and counting by analysing satellite imagery overcomes this management hurdle however, high resolution satellite imagery is expensive. Thus, to minimise cost the lowest spatial resolution data that enables accurate livestock detection should be selected. In our work, we determine the association between object detector performance and spatial degradation for cattle, sheep and dogs. Accurate ground truth was established using high resolution drone images which were then downsampled to various ground sample distances (GSDs). Both circular and cassegrain aperture optics were simulated to generate point spread functions (PSFs) corresponding to various optical qualities. By simulating the PSF, rather than approximating it as a Gaussian, the images were accurately degraded to match the spatial resolution and blurring structure of satellite imagery. 

        Two existing datasets were combined and used to train and test a YoloV5 object detection network. Detector performance was found to drop steeply around a GSD of 0.5m/px and was associated with PSF matrix structure within this GSD region. Detector mAP performance fell by 52\% when a cassegrain, rather than circular, aperture was used at a 0.5m/px GSD. Overall blurring magnitude also had a small impact when matched to GSD, as did the internal network resolution. Our results here inform the selection of remote sensing data requirements for animal detection tasks, allowing farmers and ecologists to use more accessible medium resolution imagery with confidence. 
	\end{abstract}
\end{frontmatter}
	

\section{Introduction}
Automated livestock counting and localisation from remote platforms is becoming an important tool for farmers to understand grazing patterns, track cattle health, and inform paddock management strategies \citep{barbedo2019,shao2020,wang2020}. On large Australian beef cattle stations, the cost of mustering is a significant factor in productive efficiency, complicated by the huge geographic ranges of cattle. Aircraft are used for counting and mustering, but fuel and pilot time are wasted searching for animals or herds over large areas. First locating cattle using satellites or high altitude drones could direct mustering efforts, saving time and money, while improving pilot safety. This requires localising individual cattle in a diversity of environments. The same challenges and benefits also apply to ecological monitoring of large wild animals as highlighted by \cite{weinstein2018}. 

Within agriculture and ecological science, many automated detection, counting, and tracking approaches have been demonstrated using remotely sensed or aerial images, as in \citep{xue2017,rivas2018,sarwar2018,wang2019,barbedo2020b,laradji2020}. Large individuals, or groups of animals, have been tracked using satellite sensors, while the higher resolution of aerial platforms enables small animal tracking. Across all platforms and sensing modalities, there is a trade-off between coverage area, resolution, cost, and acquisition time. While most livestock animals are of sufficient size to be distinguished in the highest resolution commercial satellite imagery, currently limited to 25cm per pixel by government regulations, this imagery is not economical to frequently gather for farms. However, as the cost of satellite construction, launch, and data acquisition falls, it will become possible to regularly image entire farms from space to locate and monitor livestock. 

By characterising the relationship between image spatial quality and animal detectability, two relevant questions can be answered. First, what resolution of imagery is required for a task needing a certain localisation and counting accuracy. Second, what detection accuracy can be expected for a given remote sensing data source. Answering the first of these can save farmers money and time when selecting remote sensing data providers, while the second has implications for ecological studies that use autonomous animal detection. 

In this paper, a large dataset of annotated, high resolution aerial farm images was modified by spatial degradation to simulate various resolutions of aerial and satellite imagery. Image sensor resolution, as measured by ground sample distance (GSD), was simulated using bicubic downsampling. Optical resolution, as captured by the point spread function (PSF), was estimated directly from the aperture function. We numerically simulated the PSF for two common aperture types and 3 aperture diameters, across all GSD values. A state of the art object detector architecture was applied to these various resolution datasets to precisely characterise the empirical relationship between spatial resolution, and counting \& localisation accuracy. 

While several research works, such as \cite{shao2020}, have approached the issue of detectability by resolution, no comprehensive study exists for animals of cattle size across a wide range of both detector and optical resolution conditions. Where optical blur is considered, existing literature uses Gaussian approximations which are unable to capture complex aperture shapes. In this study the optical point spread function was directly estimated, allowing for more realistic spatial degradation simulations. The results here can inform selection of remote sensing imagery for cow sized livestock and wild animal autonomous localisation \& counting.


\section{Related Work}

Animal detection from drones, crewed aerial platforms, and satellites is a well established technique, with many automated approaches to this proposed in the literature \citep{rivas2018,weinstein2018,kellenberger2018,chabot2018}. \cite{wang2019} provide a thorough review of remote sensing platforms and data types for wild animal surveying. Animals over 0.6m in size can be well detected in sub-meter satellite imagery, and the presence of smaller animals is often visible by environmental markers, such as bird guano. By aggregating studies, a rough relationship between spatial resolution and detectable animal size can be established. However each study uses a different, often manual, detection approach with unreliable ground truth. \cite{hollings2018} also criticise the low or unknown accuracy of remote sensing ecological survey papers, their limited test-data domains and cost of high resolution data.

A convolutional neural network (CNN) is used for cattle detection and counting by \cite{shao2020} who apply this to downwards looking unmanned aerial vehicle (UAV) imagery. With a known drone height and approximate animal size they are able to resize the CNN input resolution for improved detection performance. Testing on the older YoloV2 network architecture showed that moderate size images perform best, though no optical degradation is added to these. 

By extracting small image tiles, \cite{barbedo2019} are able to use a deep learning classifier, rather than detector, to divide tiles into cow or non-cow classes. They apply image degradation to explore optimal GSD for downwards looking UAV images, but only consider GSDs of 1, 2 and 4cm/pixel, corresponding to drone altitudes of 30, 60 and 120m. No optical blurring is added, meaning the degraded image will be unrealistically sharp due to pixel sub sampling. \cite{barbedo2020b} use oblique UAV images which implicitly vary the GSD of cattle, but do not degrade optical resolution to match the GSD. 

Many works focus on a specific aspect of cattle detection in UAV imagery. \cite{barbedo2020} improve performance on clustered animals under contrast changes by combining a deep learning detector with color space manipulation and morphology operations. Video data from a UAV is used with a region proposal type CNN to count sheep by \cite{sarwar2018}. Wide scale cattle detection in 31cm pansharpened satellite imagery is applied by \cite{laradji2020} to track illegal ranching.

Animal detection at lower resolutions has been accomplished by looking for environmental features highly correlated with the target animal presence, or by applying change detection to remove static objects. Collective wombat burrows have been identified in LANDSAT imagery as early as 1980 by \cite{loffler1980}. Penguin colonies are a common target because guano, plumage and other waste darkens the surrounding snow in multiple wavelengths \citep{schwaller1989,guinet1995,witharana2016}. Multiple images of the same area are used by \cite{stapleton2014} to distinguish polar bears from terrain features. Super resolution can also improve detection results in medium resolution images, as shown by \cite{shermeyer2019}. 

Both commercial and research tools for optical simulation exist \citep{coppo2013,fiete2014}. However, the task of degrading imagery to emulate a different remote sensing platform is uncommon in the literature. The available commercial tools did not meet the needs of this study and were not used.

Following cattle localisation, ground based or low altitude platforms can be tasked for individual animal association, body condition scoring, live-weight estimation and mustering \citep{qiao2019,qiao2020,qiao2021}. The present study is the first to explicitly examine the relationship between spatial resolution and detectability for cattle sized animals, while employing an accurate optical degradation method. 

\section{Method}
To test the impacts of spatial degradation on farm animal detectability and counting, two existing aerial farm imagery datasets were combined, degraded and used to train an object detector network. A large existing cattle dataset from \cite{shao2020} was combined with our own labelled drone survey imagery of cows, sheep and dogs. These aerial images were spatially degraded using a simulated optical point spread function and pixel downsampling to emulate data gathered from various resolution aerial or satellite sensors. Both a circular and cassegrain-style aperture PSF were simulated. Labelled animal locations are preserved from the high resolution data and were used to train a YoloV5 object detector at each reduced resolution. Finally, these trained object detection networks were assessed for their localisation and counting accuracy.

\subsection{Datasets}
The primary focus of this work was cattle detection, but identifying other common farm animals is also beneficial. So the 670 images with 1948 total cow annotations from \cite{shao2020} were combined with 475 of our own images. These additional images contain 4652 cattle, 2069 sheep, and 204 dog annotations. The combined dataset covers a wide range of terrains including grass, shrubs, dense trees, water points and low-feed paddocks. 

The total of 1145 images were randomly split into sets of 342 test, 698 train, and 114 validation images. Both of these datasets were gathered using DJI drones with downwards facing cameras, flying between 40-50m altitude. By using automatically calibrated DJI camera sensors, the optical properties of the high resolution images in both datasets are similar. Both datasets have been annotated with bounding boxes for ground truth animal locations and counts. 

\subsection{GSD and PSF Calculation}
\label{sec:Q}
Image degradation occurs at all stages in the imaging chain, from motion blur due to relative target movement, to sensor shot noise and data compression artefacts. However, two sources of spatial degradation dominate image quality reduction in remote sensing data; sensor resolution and optical blurring. 

Sensor resolution, as measured by GSD, is limited by the distance between sensor pixels, known as the pixel pitch. Ground sample distance is defined using

\begin{equation}
    GSD = \frac{pA} {f}
\end{equation}

where p is the sensor pixel pitch in $m/pix$, A is the platform altitude and $f$ is the focal distance. To simplify the calculations, a single A value was assumed across the image plane, despite the earth not being flat.

Optical diffraction blurring occurs due to light diffraction at the aperture and is captured in the point spread function of the lens. If the scene is modelled as a collection of radiating point sources $f(x,y)$, then each image is impacted by diffraction induced optical blur according to 

\begin{equation}
    I(x,y) = PSF(x,y) \circledast f(x,y)
\end{equation}

where each point in the image I, is a convolution of the point spread function with the scene points $f(x,y)$. Most optical models assume a shift invariant system, meaning the PSF is constant with regards to the scene point location, and $PSF(x,y) = PSF$.

The PSF of simple aperture geometries can be expressed analytically using Huygen's principal and the Fresnel-Kirchoff equation to model wave propagation through an aperture. Under the assumption of far-field diffraction (Fraunhofer diffraction), the light field strength at a point is approximated by \cite{fiete2010} as 

\begin{equation}
    E(x, y) \approx \frac{E_{0} e^{i k e} e^{i k\left[\frac{x^{2}+y^{2}}{2 z}\right]}}{i \lambda z} F T\left\{a\left(x_{0}, y_{0}\right)\right\}_{x_{0}=\frac{x}{\lambda z}, y_{0}=\frac{y}{\lambda z}}
    \label{eqn:waveEquation}
\end{equation}

where $E_0$ is the incoming wave source distribution, FT is the Fourier transform, $a(\cdot)$ is the aperture function, \{x,y,z\} are scene coordinates and k is the wavenumber. $\lambda$ is the radiation wavelength and $\lambda = 550$nm was used throughout.

\begin{figure*}[h!]
	\centering	
	\includegraphics[width=0.7\textwidth]{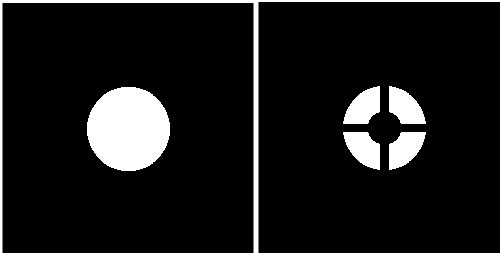}
	\caption{Examples of the circular (left) and cassegrain type (right) aperture functions used}
	\label{fig:apertures}
\end{figure*}

\begin{figure*}[h!]
	\centering	
	\includegraphics[width=0.8\textwidth]{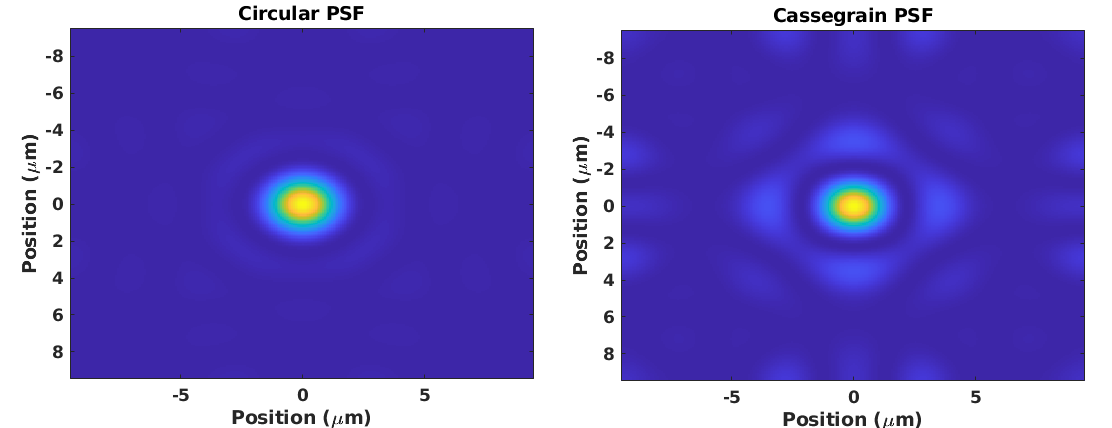}
	\caption{An example of a calculated circular (left) and cassegrain (right) point spread function}
	\label{fig:PSFs}
\end{figure*}

For a circular aperture of diameter D, the aperture function for any point at radius r relative to the circle centre is given by

\begin{equation}
    \operatorname{circ}\left(\frac{r}{D}\right)= \begin{cases}1, & r<\frac{D}{2} \\ \frac{1}{2} & r=\frac{D}{2} \\ 0, & r>\frac{D}{2} .\end{cases}
    \label{eqn:circularApertureEquation}
\end{equation}

and Equation \ref{eqn:waveEquation} becomes

\begin{equation}
    PSF_{circularAperture} = \left(\frac{E_{0} \pi D^{2}}{4 \lambda f}\right)^{2}\left|\frac{2 J_{1}\left(\frac{\pi D r}{\lambda f}\right)}{\frac{\pi D r}{\lambda f}}\right|^{2}
    \label{eqn:circularDiffractionEquation}
\end{equation}

where $J_1$ is the first order Bessel function. One convenient measure of the PSF spread is the diameter of the first zero of the Airy disk, which occurs at $2.44 \lambda f/D$. A common technique is to approximate the PSF using a 2D Gaussian of equal first zero size, but this reduces accuracy and cannot capture non-circular aperture functions. Instead, we numerically simulated the PSF by taking the fast Fourier transform of the aperture function. Examples of this are shown in Figures \ref{fig:apertures} and \ref{fig:PSFs}.

The Nyquist criteria can be used to determine the minimum size feature resolvable under the optical diffraction limits. If the smallest optically resolvable feature is below the ground sample distance, the system is said to be \textit{detector limited}. In the converse case it is \textit{diffraction limited}. This notion can be formalised in the Q value, defined by

\begin{equation}
    Q = 2 \frac{resolution_{optical}}{resolution_{detector}} = \frac{\lambda f}{D p}
\end{equation}

Selection of Q for actual imaging systems is complex and driven by impacts beyond spatial sampling resolution. Most cameras operate in the region $0.5 < Q < 1.5$ so this spread of values was used when simulating imagery. For comparison, the Planet SkySat-C constellation provides 0.7m to 0.9m GSD imagery using a Q value of 0.87, as calculated from values in \cite{bhushan2021}.  

\subsection{Image Degradation Model}
The spatial resolution degradation model, used to simulate satellite imagery from high resolution drone data, consisted of image blurring followed by decimation. This used bicubic-downsampling, and is defined as

\begin{equation}
I_{sat} = (I_{uav} \circledast PSF)\downarrow _\phi
\label{eqn:degradation}
\end{equation}

where $I_{sat}$ is the simulated satellite image, $I_{uav}$ is the source UAV image, $\circledast$ is the convolution operator and $\downarrow _\phi$ denotes a pixel resampling such that 

\begin{equation}
    GSD_{sat} = \phi\times GSD_{uav}
\end{equation}

For clarity low and high resolution imagery is described as satellite and UAV data respectively, but a higher altitude drone or crewed aircraft is equivalently simulated using this general degradation model. \cite{zhang2020} point out that the correct choice of blurring kernel in Equation \ref{eqn:degradation} can approximate the bicubic resampling degradation method, though separating these steps simplifies the PSF calculation. Bicubic decimation is a common and effective pixel resampling method. It can also be used to upscale imagery, where the additional pixel values are bicubically interpolated.

Only the pixel resampling operation directly impacts GSD of the data, so this was set to produce the desired GSD values. The blurring kernel would normally capture the imaging system modulation transfer function (MTF), which includes additional optical effects, but we seek platform agnostic results, and known MTFs can be largely corrected for. So instead the blur kernel was only used to capture diffraction limit induced blur, defined by the PSF. We assume here that $\phi$ is sufficiently large that the MTF spatial extents will fall within a single downsampled pixel, resulting in no blurring contribution from the UAV MTF in the downsampled imagery. 

Both a circular aperture function and that of a cassegrain type telescope were tested, as shown in Figure \ref{fig:apertures}. Approximately circular apertures are found on most cameras and many satellite types, and produce a circularly symmetrical PSF. Cassegrain type telescopes use a secondary mirror which folds the optical path, and are common in larger satellites or aerial platforms. The second mirror and its supports obscure portions of the image, resulting in an irregular PSF. The cassegrain aperture was constructed by masking the circular aperture where the central mirror and supports would be, so it has an equal diameter but smaller overall area. Figure \ref{fig:PSFs} shows an example of each PSF type. This discrete PSF kernel was calculated for each Q value, and resized to maintain this Q value at the pixel plane when the change in image GSD was applied by bicubic resampling.

\begin{figure*}[ht]
	\centering	
	\includegraphics[width=0.95\textwidth]{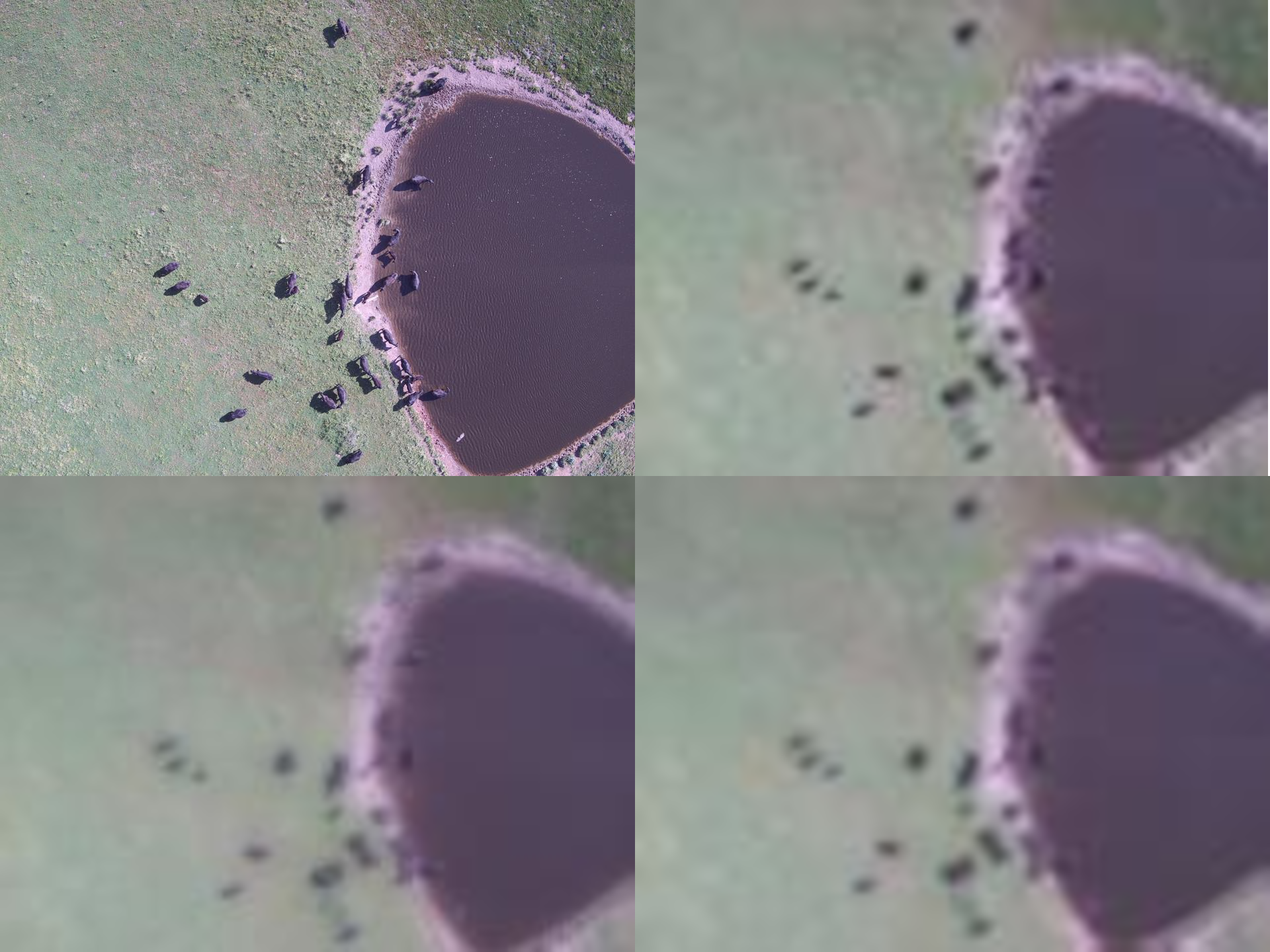}
	\caption{Example images showing (clockwise from top left) a 0.05m GSD image at Q=0.5 with circular PSF, a 0.5m GSD image at Q=0.5 with circular PSF, a 0.5m GSD image at Q=1.5 with circular PSF, and a 0.5m GSD image at Q=1.5 with cassegrain PSF.}
	\label{fig:dataExample}
\end{figure*}

Convolving this PSF kernel with large images is slow, so a two stage process was used. The input image was downsampled to an intermediate size which was chosen to result in a PSF kernel of dimensions 64x64, which is large enough to accurately capture the PSF shape. The PSF kernel was convolved with this lower resolution image using a sliding window approach, before being further bicubically downsampled to the final GSD target. Figure \ref{fig:dataExample} shows images degraded using this method. 

\subsection{Detection \& Counting Method}
Automatic counting methods include detection based, density map based and direct regression from image frames. Bounding boxes provide individual animal locations for farmers, and are the most common format for object detectors, so were used. 

YoloV5 was chosen as a current generation object detection framework and we used the most commonly recognised implementation\footnote{https://github.com/ultralytics/yolov5}. YoloV5 internally resizes images using a selectable input size parameter, this was tested at 640 and 1280 pixels. Network hyperparameters were tuned using a genetic algorithm applied to the full resolution imagery. The best fit parameters were found after 541 generations of 10 epochs each and a batch size of 4. Genetic evolution hyperparameter tuning could be applied to each resolution but is very computationally expensive, at several GPU weeks per solution.

\section{Results}
Average precision (AP), mean average precision (mAP) across classes and the precision-recall harmonic mean (F1) score are used to quantify object detector performance and localisation. All values are reported for an IoU threshold of 0.5. To measure count performance, the mean absolute count error (CE) per image is defined as

\begin{equation}
    CE = \frac{1}{N}\sum_{n=1}^{N} |D_n - L_n|
\end{equation}

where N is the number of images in the test set, $D_n$ is the number of predicted detections in image n above a given confidence threshold and $L_n$ is the number of bounding box labels in that image. The detector confidence threshold was determined per dataset as the value which maximises the F1 score. The model trained at full resolution was also run on the lower resolution test sets, but performance was much worse than models trained at these lower resolutions. 

\subsection{Performance By Aperture \& Input Resolution}
\begin{figure*}[h!]
	\centering	
	\includegraphics[width=0.7\textwidth]{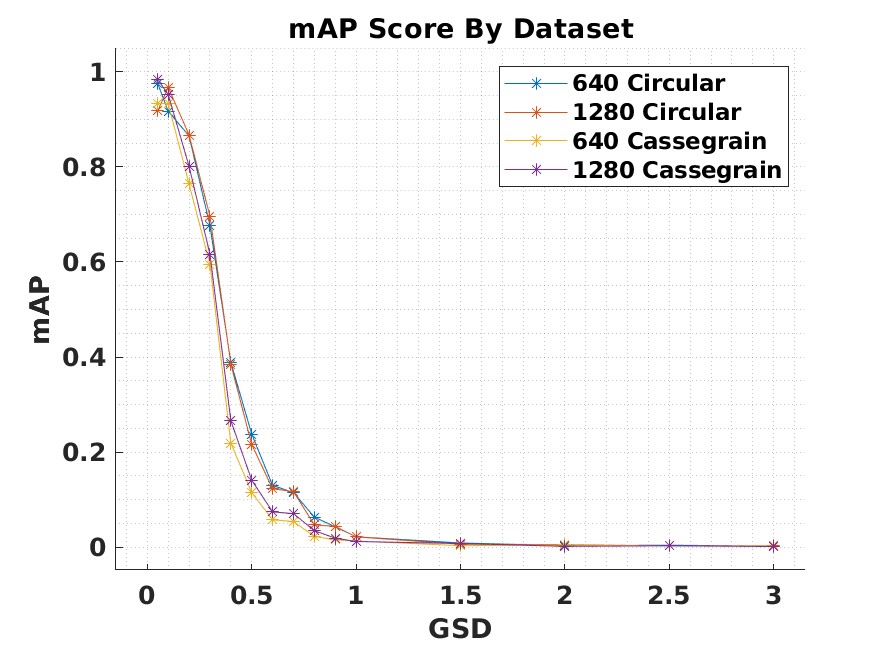}
	\caption{Comparison of mAP performance by resolution for each dataset at Q = 1.0}
	\label{fig:mAP}
\end{figure*}

\begin{figure*}[h!]
	\centering	
	\includegraphics[width=0.7\textwidth]{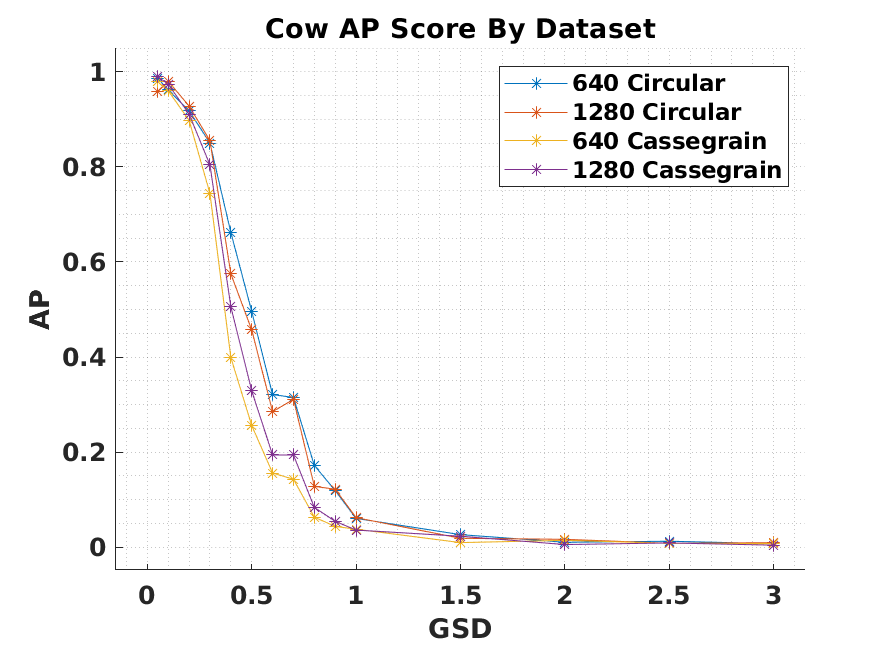}
	\caption{Comparison of average precision for the cow class, by resolution, for each dataset at Q = 1.0}
	\label{fig:cowAP}
\end{figure*}

Results in this section are for a Q value of 1.0, and consider the impact of aperture shape and YoloV5 network input resolution. It is clear from Figures \ref{fig:mAP} and \ref{fig:cowAP} that the additional magnitude and structure of blurring induced by the cassegrain style aperture function reduced object detectability. This reduction was strongest at GSD values around 0.5m/px, which is where detector performance was most sensitive to spatial degradation changes. This is also the region where systems should currently operate if trying to minimise the camera requirements for animal detection using remote sensing. Because of this, optical quality and aperture shape may play an outsized role for systems seeking to use the lowest possible GSD for detection. 

\begin{figure*}[h!]
	\centering	
	\includegraphics[width=0.95\textwidth]{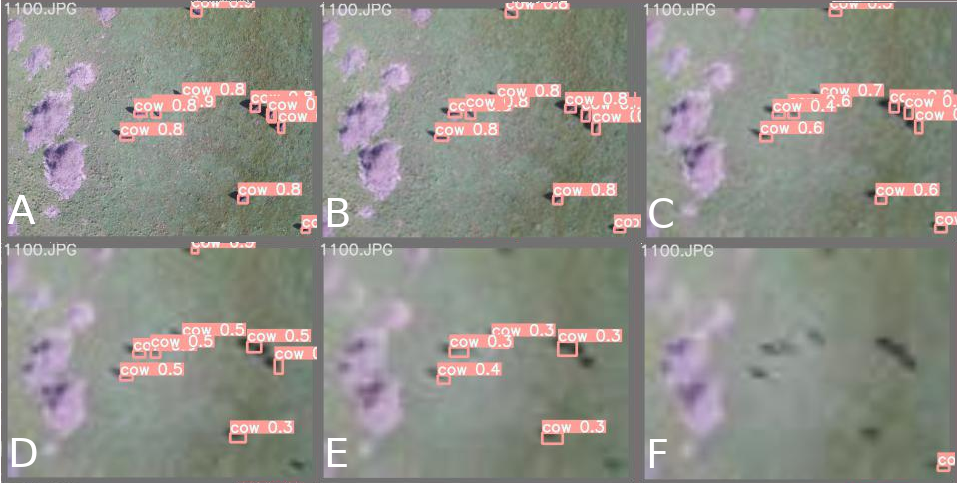}
	\caption{Example detection results on one of the validation set images for a circular aperture with 640px input size and GSD of: a) 0.05m, b) 0.2m, c) 0.4m, d) 0.6m, e) 0.8m, f) 1.0m}
	\label{fig:resultsExample}
\end{figure*}

\begin{figure*}[h!]
	\centering	
	\includegraphics[width=0.7\textwidth]{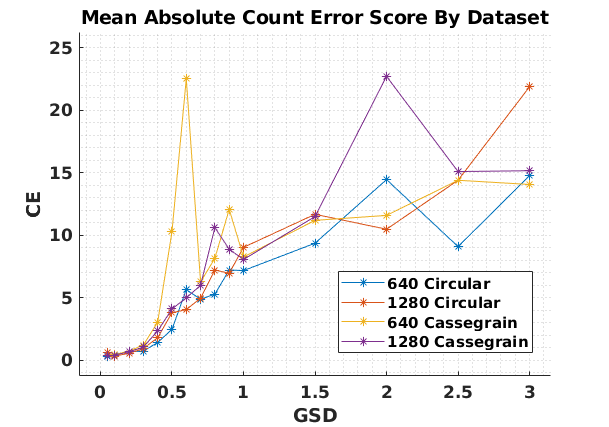}
	\caption{Comparison of the count error performance by resolution for each dataset}
	\label{fig:CE}
\end{figure*}

Unlike in \cite{shao2020}, network input size was found to have a small impact on detector performance. Degraded images produced for every GSD are below 1280px in size, so network input sizes resulted in a bicubic upsampling for all data. The lack of performance boost when using a larger network input size suggests that animals being below the minimum feature size or smallest anchor proposal size for YoloV5 is not a significant problem. This indicates that lack of image detail, rather than network parameters or settings, is limiting the detectability of animals of this size for YoloV5. 

Count error showed the same trend of a circular aperture performing best, with a slight benefit to using higher network input resolution. The mean number of objects per image is 3.4, and above approximately 0.5m/px the count values became meaningless. Examination of the count error histograms for various GSD values above 0.5m/px shows the error is roughly Gaussian, so no correction could be applied to improve the accuracy of systematic over-counting or under-counting.

Cows followed the same trend as other animal classes, but with better detectability across the board. This is likely due to their larger size, and because many of the cows are black in colour and more highly contrast against the background terrain than white sheep or brown dogs.

\subsection{Performance By Q Value}
\begin{figure*}[h!]
	\centering	
	\includegraphics[width=0.7\textwidth]{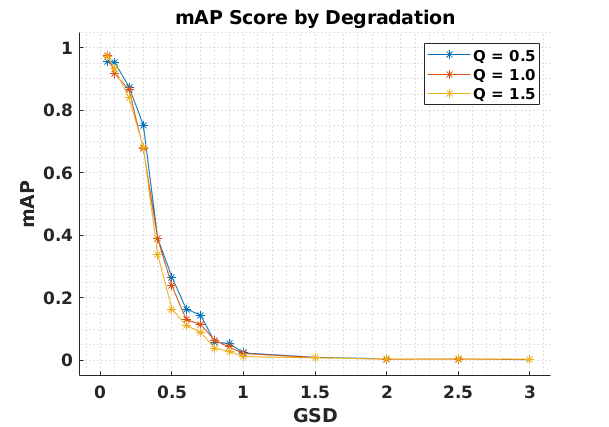}
	\caption{mAP performance on the 640 circular dataset by GSD and Q value}
	\label{fig:circularmAP}
\end{figure*}

\begin{figure*}[h!]
	\centering	
	\includegraphics[width=0.7\textwidth]{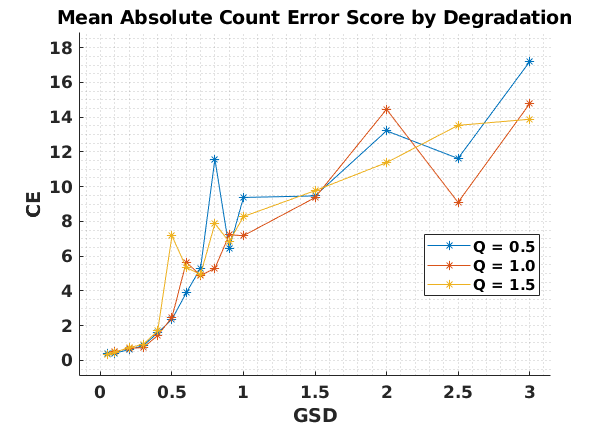}
	\caption{Mean absolute count error performance on the 640 circular dataset by GSD and Q value}
	\label{fig:circularCE}
\end{figure*}

The data in this section is for the circular aperture dataset at 640px YoloV5 input resolution. Unlike aperture shape, no clear link was found between Q value and count error, as shown in Figure \ref{fig:circularCE}. So optical blurring magnitude alone is less of a contributor to count error, than structure in the PSF. The mAP across all classes was impacted by Q, with higher optical quality performing better, as shown in Figure \ref{fig:circularmAP}. Although this effect was much smaller than GSD changes.

\section{Conclusions}
As low cost satellite data becomes more available, the opportunities to use automated animal localisation and counting to benefit farmers and ecologists grow. Our results showed that 0.5m/px GSD is the threshold required for this technique to be applied to cattle sized animals, depending on accuracy requirements. 

Cassegrain aperture induced PSF structure, played a minimal role at very small ground sample distances. However, for targets on the edge of detectability due to GSD, this was much more influential and should be closely considered. Varying Q directly, by reducing the aperture diameter, had a noticeable, but much smaller impact. Ground sample distance had clear implications for object detector performance in remote sensing, with performance falling rapidly as GSD moved from 0.2m/px to 0.5m/px. While there are many techniques to improve this at a given GSD, the overall trend will dictate remote sensing requirements for animal localisation and counting. 

Obscuration by clouds and dense tree cover remains an issue, and was not addressed by this study. Techniques for overcoming this, such as synthetic aperture radar or infrared imaging, may have similar detectability by resolution trends as were found here. Future studies are also required to better understand how detectability is impacted by other aspects of the remote sensing image chain, including atmospheric turbulence blurring, chromatic aberration and compression artefacts. Using bi-temporal or multi-temporal images to spot moving objects can help distinguish rocks from cattle. We are currently investigating this, and other methods, for improving detectability results around the 0.5m/px GSD range. 

\section*{ACKNOWLEDGMENTS}
The authors acknowledge the support of the Meat \& Livestock Australia Donor Company through the project: Objective, robust, real-time animal welfare measures for the Australian red meat industry (P.PSH.0819). 

\section*{CRediT authorship contribution statement}
J.B..: Conceptualization, Investigation, Data curation, Methodology, Validation, Formal analysis, and Writing-original draft; Y.Q.: Data gathering, Data formatting, and Writing - review \& editing; C.C: Funding acquisition, Writing - review \& editing; S.L: Writing - review \& editing; K.R: Resources, Writing - review \& editing; S.S.: Resources, Writing - review \& editing.

\section*{Declaration of competing interest}
The authors declare that they have no known competing financial interests or personal relationships that could have appeared to influence the work reported in this paper.

\newpage
\bibliography{DetectabilityByGSD}
\bibliographystyle{apalike}

\newpage
\section{Appendix: Numerical Results}
\input{table_standard}
\input{table_highres}
\input{table_cass}
\input{table_cassHighres}
\end{document}

%% file: table_standard.tex
\begin{table}[h!]
\resizebox{0.95\textwidth}{!}{
\begin{tabular}{|c|c|c|c|c|c|c|c|c|c|}
\hline
\textbf{\begin{tabular}[c]{@{}c@{}}GSD \\ (m/px)\end{tabular}} & \textbf{Q}            & \textbf{F1} & \textbf{\begin{tabular}[c]{@{}c@{}}Mean Abs\\ Count Error\end{tabular}} & \textbf{\begin{tabular}[c]{@{}c@{}}Precision\\ All Classes\end{tabular}} & \textbf{\begin{tabular}[c]{@{}c@{}}Recall\\ All Classes\end{tabular}} & \textbf{\begin{tabular}[c]{@{}c@{}}mAP\\ All Classes\end{tabular}} & \textbf{\begin{tabular}[c]{@{}c@{}}AP\\ Cow\end{tabular}} & \textbf{\begin{tabular}[c]{@{}c@{}}AP\\ Sheep\end{tabular}} & \textbf{\begin{tabular}[c]{@{}c@{}}AP\\ Dog\end{tabular}} \\ \hline
0.05                                                           & \multirow{15}{*}{0.5} & 0.93        & 0.375                                                                   & 0.931                                                                    & 0.937                                                                 & 0.955                                                              & 0.989                                                     & 0.988                                                       & 0.887                                                     \\ \cline{1-1} \cline{3-10} 
0.10                                                           &                       & 0.93        & 0.407                                                                   & 0.930                                                                    & 0.937                                                                 & 0.953                                                              & 0.972                                                     & 0.984                                                       & 0.902                                                     \\ \cline{1-1} \cline{3-10} 
0.20                                                           &                       & 0.88        & 0.625                                                                   & 0.902                                                                    & 0.869                                                                 & 0.874                                                              & 0.934                                                     & 0.985                                                       & 0.702                                                     \\ \cline{1-1} \cline{3-10} 
0.30                                                           &                       & 0.76        & 0.850                                                                   & 0.823                                                                    & 0.727                                                                 & 0.752                                                              & 0.871                                                     & 0.970                                                       & 0.416                                                     \\ \cline{1-1} \cline{3-10} 
0.40                                                           &                       & 0.44        & 1.585                                                                   & 0.527                                                                    & 0.397                                                                 & 0.390                                                              & 0.622                                                     & 0.491                                                       & 0.058                                                     \\ \cline{1-1} \cline{3-10} 
0.50                                                           &                       & 0.3         & 2.365                                                                   & 0.331                                                                    & 0.289                                                                 & 0.266                                                              & 0.563                                                     & 0.229                                                       & 0.007                                                     \\ \cline{1-1} \cline{3-10} 
0.60                                                           &                       & 0.21        & 3.886                                                                   & 0.232                                                                    & 0.214                                                                 & 0.164                                                              & 0.382                                                     & 0.096                                                       & 0.013                                                     \\ \cline{1-1} \cline{3-10} 
0.70                                                           &                       & 0.2         & 5.310                                                                   & 0.235                                                                    & 0.194                                                                 & 0.145                                                              & 0.354                                                     & 0.076                                                       & 0.005                                                     \\ \cline{1-1} \cline{3-10} 
0.80                                                           &                       & 0.09        & 11.585                                                                  & 0.104                                                                    & 0.128                                                                 & 0.058                                                              & 0.157                                                     & 0.015                                                       & 0.001                                                     \\ \cline{1-1} \cline{3-10} 
0.90                                                           &                       & 0.08        & 6.436                                                                   & 0.750                                                                    & 0.080                                                                 & 0.054                                                              & 0.153                                                     & 0.010                                                       & 0.001                                                     \\ \cline{1-1} \cline{3-10} 
1.00                                                           &                       & 0.05        & 9.377                                                                   & 0.709                                                                    & 0.066                                                                 & 0.024                                                              & 0.065                                                     & 0.006                                                       & 0.000                                                     \\ \cline{1-1} \cline{3-10} 
1.50                                                           &                       & 0.03        & 9.465                                                                   & 0.703                                                                    & 0.027                                                                 & 0.010                                                              & 0.029                                                     & 0.000                                                       & 0.000                                                     \\ \cline{1-1} \cline{3-10} 
2.00                                                           &                       & 0.03        & 13.211                                                                  & 0.690                                                                    & 0.031                                                                 & 0.005                                                              & 0.015                                                     & 0.000                                                       & 0.000                                                     \\ \cline{1-1} \cline{3-10} 
2.50                                                           &                       & 0.02        & 11.622                                                                  & 0.362                                                                    & 0.022                                                                 & 0.005                                                              & 0.014                                                     & 0.000                                                       & 0.000                                                     \\ \cline{1-1} \cline{3-10} 
3.00                                                           &                       & 0.02        & 17.184                                                                  & 0.346                                                                    & 0.027                                                                 & 0.002                                                              & 0.006                                                     & 0.000                                                       & 0.000                                                     \\ \hline
0.05                                                           & \multirow{15}{*}{1.0} & 0.96        & 0.344                                                                   & 0.958                                                                    & 0.955                                                                 & 0.975                                                              & 0.986                                                     & 0.985                                                       & 0.954                                                     \\ \cline{1-1} \cline{3-10} 
0.10                                                           &                       & 0.91        & 0.506                                                                   & 0.901                                                                    & 0.910                                                                 & 0.917                                                              & 0.962                                                     & 0.984                                                       & 0.805                                                     \\ \cline{1-1} \cline{3-10} 
0.20                                                           &                       & 0.85        & 0.689                                                                   & 0.855                                                                    & 0.854                                                                 & 0.865                                                              & 0.918                                                     & 0.985                                                       & 0.692                                                     \\ \cline{1-1} \cline{3-10} 
0.30                                                           &                       & 0.71        & 0.738                                                                   & 0.745                                                                    & 0.684                                                                 & 0.676                                                              & 0.850                                                     & 0.952                                                       & 0.226                                                     \\ \cline{1-1} \cline{3-10} 
0.40                                                           &                       & 0.43        & 1.430                                                                   & 0.437                                                                    & 0.426                                                                 & 0.389                                                              & 0.662                                                     & 0.476                                                       & 0.029                                                     \\ \cline{1-1} \cline{3-10} 
0.50                                                           &                       & 0.28        & 2.477                                                                   & 0.297                                                                    & 0.264                                                                 & 0.238                                                              & 0.496                                                     & 0.208                                                       & 0.009                                                     \\ \cline{1-1} \cline{3-10} 
0.60                                                           &                       & 0.18        & 5.655                                                                   & 0.208                                                                    & 0.191                                                                 & 0.130                                                              & 0.321                                                     & 0.061                                                       & 0.006                                                     \\ \cline{1-1} \cline{3-10} 
0.70                                                           &                       & 0.14        & 4.883                                                                   & 0.187                                                                    & 0.147                                                                 & 0.115                                                              & 0.315                                                     & 0.029                                                       & 0.002                                                     \\ \cline{1-1} \cline{3-10} 
0.80                                                           &                       & 0.09        & 5.295                                                                   & 0.777                                                                    & 0.070                                                                 & 0.063                                                              & 0.172                                                     & 0.016                                                       & 0.002                                                     \\ \cline{1-1} \cline{3-10} 
0.90                                                           &                       & 0.07        & 7.222                                                                   & 0.733                                                                    & 0.076                                                                 & 0.043                                                              & 0.119                                                     & 0.007                                                       & 0.003                                                     \\ \cline{1-1} \cline{3-10} 
1.00                                                           &                       & 0.05        & 7.178                                                                   & 0.717                                                                    & 0.055                                                                 & 0.022                                                              & 0.061                                                     & 0.005                                                       & 0.000                                                     \\ \cline{1-1} \cline{3-10} 
1.50                                                           &                       & 0.03        & 9.365                                                                   & 0.706                                                                    & 0.025                                                                 & 0.009                                                              & 0.027                                                     & 0.000                                                       & 0.000                                                     \\ \cline{1-1} \cline{3-10} 
2.00                                                           &                       & 0.02        & 14.449                                                                  & 0.684                                                                    & 0.025                                                                 & 0.003                                                              & 0.010                                                     & 0.000                                                       & 0.000                                                     \\ \cline{1-1} \cline{3-10} 
2.50                                                           &                       & 0.02        & 9.094                                                                   & 0.356                                                                    & 0.014                                                                 & 0.004                                                              & 0.013                                                     & 0.000                                                       & 0.000                                                     \\ \cline{1-1} \cline{3-10} 
3.00                                                           &                       & 0.02        & 14.763                                                                  & 0.351                                                                    & 0.025                                                                 & 0.002                                                              & 0.007                                                     & 0.000                                                       & 0.000                                                     \\ \hline
0.05                                                           & \multirow{15}{*}{1.5} & 0.96        & 0.345                                                                   & 0.947                                                                    & 0.967                                                                 & 0.972                                                              & 0.984                                                     & 0.983                                                       & 0.948                                                     \\ \cline{1-1} \cline{3-10} 
0.10                                                           &                       & 0.92        & 0.469                                                                   & 0.917                                                                    & 0.923                                                                 & 0.932                                                              & 0.963                                                     & 0.983                                                       & 0.850                                                     \\ \cline{1-1} \cline{3-10} 
0.20                                                           &                       & 0.86        & 0.727                                                                   & 0.885                                                                    & 0.844                                                                 & 0.839                                                              & 0.916                                                     & 0.980                                                       & 0.622                                                     \\ \cline{1-1} \cline{3-10} 
0.30                                                           &                       & 0.71        & 0.935                                                                   & 0.796                                                                    & 0.660                                                                 & 0.681                                                              & 0.846                                                     & 0.931                                                       & 0.267                                                     \\ \cline{1-1} \cline{3-10} 
0.40                                                           &                       & 0.39        & 1.699                                                                   & 0.421                                                                    & 0.371                                                                 & 0.338                                                              & 0.616                                                     & 0.355                                                       & 0.044                                                     \\ \cline{1-1} \cline{3-10} 
0.50                                                           &                       & 0.19        & 7.170                                                                   & 0.162                                                                    & 0.254                                                                 & 0.164                                                              & 0.397                                                     & 0.093                                                       & 0.003                                                     \\ \cline{1-1} \cline{3-10} 
0.60                                                           &                       & 0.14        & 5.339                                                                   & 0.204                                                                    & 0.150                                                                 & 0.112                                                              & 0.287                                                     & 0.046                                                       & 0.002                                                     \\ \cline{1-1} \cline{3-10} 
0.70                                                           &                       & 0.11        & 4.942                                                                   & 0.476                                                                    & 0.095                                                                 & 0.090                                                              & 0.254                                                     & 0.013                                                       & 0.003                                                     \\ \cline{1-1} \cline{3-10} 
0.80                                                           &                       & 0.07        & 7.871                                                                   & 0.727                                                                    & 0.075                                                                 & 0.039                                                              & 0.106                                                     & 0.006                                                       & 0.005                                                     \\ \cline{1-1} \cline{3-10} 
0.90                                                           &                       & 0.06        & 6.827                                                                   & 0.731                                                                    & 0.061                                                                 & 0.029                                                              & 0.079                                                     & 0.007                                                       & 0.001                                                     \\ \cline{1-1} \cline{3-10} 
1.00                                                           &                       & 0.04        & 8.278                                                                   & 0.701                                                                    & 0.038                                                                 & 0.012                                                              & 0.035                                                     & 0.001                                                       & 0.000                                                     \\ \cline{1-1} \cline{3-10} 
1.50                                                           &                       & 0.03        & 9.766                                                                   & 0.696                                                                    & 0.026                                                                 & 0.008                                                              & 0.023                                                     & 0.000                                                       & 0.000                                                     \\ \cline{1-1} \cline{3-10} 
2.00                                                           &                       & 0.02        & 11.382                                                                  & 0.353                                                                    & 0.018                                                                 & 0.005                                                              & 0.014                                                     & 0.000                                                       & 0.000                                                     \\ \cline{1-1} \cline{3-10} 
2.50                                                           &                       & 0.02        & 13.526                                                                  & 0.354                                                                    & 0.026                                                                 & 0.005                                                              & 0.016                                                     & 0.000                                                       & 0.000                                                     \\ \cline{1-1} \cline{3-10} 
3.00                                                           &                       & 0.02        & 13.877                                                                  & 0.349                                                                    & 0.022                                                                 & 0.004                                                              & 0.011                                                     & 0.000                                                       & 0.000                                                     \\ \hline
\end{tabular}}
\caption{Object localisation and counting results for circular aperture, 640 resolution input images} 
\end{table}

%% file: table_highres.tex
\begin{table}[]
\resizebox{\textwidth}{!}{
\begin{tabular}{|l|l|l|l|l|l|l|l|l|l|}
\hline
\multicolumn{1}{|c|}{\textbf{\begin{tabular}[c]{@{}c@{}}GSD \\ (m/px)\end{tabular}}} & \multicolumn{1}{c|}{\textbf{Q}} & \multicolumn{1}{c|}{\textbf{F1}} & \multicolumn{1}{c|}{\textbf{\begin{tabular}[c]{@{}c@{}}Mean Abs\\ Count Error\end{tabular}}} & \multicolumn{1}{c|}{\textbf{\begin{tabular}[c]{@{}c@{}}Precision\\ All Classes\end{tabular}}} & \multicolumn{1}{c|}{\textbf{\begin{tabular}[c]{@{}c@{}}Recall\\ All Classes\end{tabular}}} & \multicolumn{1}{c|}{\textbf{\begin{tabular}[c]{@{}c@{}}mAP\\ All Classes\end{tabular}}} & \multicolumn{1}{c|}{\textbf{\begin{tabular}[c]{@{}c@{}}AP\\ Cow\end{tabular}}} & \multicolumn{1}{c|}{\textbf{\begin{tabular}[c]{@{}c@{}}AP\\ Sheep\end{tabular}}} & \multicolumn{1}{c|}{\textbf{\begin{tabular}[c]{@{}c@{}}AP\\ Dog\end{tabular}}} \\ \hline
0.05                                                                                 & \multirow{15}{*}{0.5}           & 0.97                             & 0.276                                                                                        & 0.963                                                                                         & 0.969                                                                                      & 0.976                                                                                   & 0.992                                                                          & 0.988                                                                            & 0.949                                                                          \\ \cline{1-1} \cline{3-10} 
0.10                                                                                 &                                 & 0.95                             & 0.336                                                                                        & 0.947                                                                                         & 0.957                                                                                      & 0.957                                                                                   & 0.978                                                                          & 0.985                                                                            & 0.907                                                                          \\ \cline{1-1} \cline{3-10} 
0.20                                                                                 &                                 & 0.87                             & 0.591                                                                                        & 0.867                                                                                         & 0.865                                                                                      & 0.892                                                                                   & 0.929                                                                          & 0.989                                                                            & 0.759                                                                          \\ \cline{1-1} \cline{3-10} 
0.30                                                                                 &                                 & 0.76                             & 0.734                                                                                        & 0.807                                                                                         & 0.730                                                                                      & 0.745                                                                                   & 0.883                                                                          & 0.975                                                                            & 0.378                                                                          \\ \cline{1-1} \cline{3-10} 
0.40                                                                                 &                                 & 0.46                             & 1.676                                                                                        & 0.529                                                                                         & 0.424                                                                                      & 0.411                                                                                   & 0.658                                                                          & 0.518                                                                            & 0.057                                                                          \\ \cline{1-1} \cline{3-10} 
0.50                                                                                 &                                 & 0.29                             & 2.456                                                                                        & 0.357                                                                                         & 0.246                                                                                      & 0.253                                                                                   & 0.464                                                                          & 0.282                                                                            & 0.014                                                                          \\ \cline{1-1} \cline{3-10} 
0.60                                                                                 &                                 & 0.21                             & 4.398                                                                                        & 0.242                                                                                         & 0.197                                                                                      & 0.150                                                                                   & 0.321                                                                          & 0.114                                                                            & 0.015                                                                          \\ \cline{1-1} \cline{3-10} 
0.70                                                                                 &                                 & 0.18                             & 3.944                                                                                        & 0.250                                                                                         & 0.162                                                                                      & 0.141                                                                                   & 0.366                                                                          & 0.052                                                                            & 0.006                                                                          \\ \cline{1-1} \cline{3-10} 
0.80                                                                                 &                                 & 0.1                              & 12.348                                                                                       & 0.088                                                                                         & 0.134                                                                                      & 0.056                                                                                   & 0.146                                                                          & 0.020                                                                            & 0.003                                                                          \\ \cline{1-1} \cline{3-10} 
0.90                                                                                 &                                 & 0.07                             & 6.541                                                                                        & 0.739                                                                                         & 0.073                                                                                      & 0.047                                                                                   & 0.131                                                                          & 0.005                                                                            & 0.005                                                                          \\ \cline{1-1} \cline{3-10} 
1.00                                                                                 &                                 & 0.05                             & 7.795                                                                                        & 0.716                                                                                         & 0.060                                                                                      & 0.027                                                                                   & 0.078                                                                          & 0.003                                                                            & 0.001                                                                          \\ \cline{1-1} \cline{3-10} 
1.50                                                                                 &                                 & 0.03                             & 9.965                                                                                        & 0.697                                                                                         & 0.040                                                                                      & 0.010                                                                                   & 0.029                                                                          & 0.000                                                                            & 0.000                                                                          \\ \cline{1-1} \cline{3-10} 
2.00                                                                                 &                                 & 0.02                             & 10.181                                                                                       & 0.688                                                                                         & 0.024                                                                                      & 0.005                                                                                   & 0.015                                                                          & 0.000                                                                            & 0.000                                                                          \\ \cline{1-1} \cline{3-10} 
2.50                                                                                 &                                 & 0.03                             & 12.129                                                                                       & 0.360                                                                                         & 0.031                                                                                      & 0.007                                                                                   & 0.022                                                                          & 0.000                                                                            & 0.000                                                                          \\ \cline{1-1} \cline{3-10} 
3.00                                                                                 &                                 & 0.01                             & 28.529                                                                                       & 0.340                                                                                         & 0.023                                                                                      & 0.001                                                                                   & 0.003                                                                          & 0.000                                                                            & 0.000                                                                          \\ \hline
0.05                                                                                 & \multirow{15}{*}{1.0}           & 0.9                              & 0.604                                                                                        & 0.928                                                                                         & 0.872                                                                                      & 0.918                                                                                   & 0.958                                                                          & 0.968                                                                            & 0.827                                                                          \\ \cline{1-1} \cline{3-10} 
0.10                                                                                 &                                 & 0.96                             & 0.342                                                                                        & 0.958                                                                                         & 0.958                                                                                      & 0.967                                                                                   & 0.980                                                                          & 0.990                                                                            & 0.932                                                                          \\ \cline{1-1} \cline{3-10} 
0.20                                                                                 &                                 & 0.87                             & 0.566                                                                                        & 0.889                                                                                         & 0.856                                                                                      & 0.867                                                                                   & 0.928                                                                          & 0.989                                                                            & 0.684                                                                          \\ \cline{1-1} \cline{3-10} 
0.30                                                                                 &                                 & 0.7                              & 0.944                                                                                        & 0.796                                                                                         & 0.656                                                                                      & 0.696                                                                                   & 0.855                                                                          & 0.966                                                                            & 0.266                                                                          \\ \cline{1-1} \cline{3-10} 
0.40                                                                                 &                                 & 0.43                             & 1.807                                                                                        & 0.561                                                                                         & 0.382                                                                                      & 0.385                                                                                   & 0.575                                                                          & 0.519                                                                            & 0.060                                                                          \\ \cline{1-1} \cline{3-10} 
0.50                                                                                 &                                 & 0.28                             & 3.807                                                                                        & 0.259                                                                                         & 0.303                                                                                      & 0.216                                                                                   & 0.457                                                                          & 0.171                                                                            & 0.019                                                                          \\ \cline{1-1} \cline{3-10} 
0.60                                                                                 &                                 & 0.17                             & 4.058                                                                                        & 0.248                                                                                         & 0.145                                                                                      & 0.123                                                                                   & 0.285                                                                          & 0.076                                                                            & 0.006                                                                          \\ \cline{1-1} \cline{3-10} 
0.70                                                                                 &                                 & 0.15                             & 4.974                                                                                        & 0.220                                                                                         & 0.143                                                                                      & 0.117                                                                                   & 0.311                                                                          & 0.033                                                                            & 0.008                                                                          \\ \cline{1-1} \cline{3-10} 
0.80                                                                                 &                                 & 0.08                             & 7.222                                                                                        & 0.736                                                                                         & 0.089                                                                                      & 0.047                                                                                   & 0.128                                                                          & 0.010                                                                            & 0.003                                                                          \\ \cline{1-1} \cline{3-10} 
0.90                                                                                 &                                 & 0.07                             & 6.962                                                                                        & 0.754                                                                                         & 0.060                                                                                      & 0.044                                                                                   & 0.122                                                                          & 0.003                                                                            & 0.007                                                                          \\ \cline{1-1} \cline{3-10} 
1.00                                                                                 &                                 & 0.05                             & 9.035                                                                                        & 0.709                                                                                         & 0.064                                                                                      & 0.022                                                                                   & 0.062                                                                          & 0.002                                                                            & 0.001                                                                          \\ \cline{1-1} \cline{3-10} 
1.50                                                                                 &                                 & 0.03                             & 11.678                                                                                       & 0.692                                                                                         & 0.032                                                                                      & 0.006                                                                                   & 0.019                                                                          & 0.000                                                                            & 0.000                                                                          \\ \cline{1-1} \cline{3-10} 
2.00                                                                                 &                                 & 0.03                             & 10.482                                                                                       & 0.695                                                                                         & 0.024                                                                                      & 0.006                                                                                   & 0.017                                                                          & 0.000                                                                            & 0.000                                                                          \\ \cline{1-1} \cline{3-10} 
2.50                                                                                 &                                 & 0.02                             & 14.397                                                                                       & 0.347                                                                                         & 0.022                                                                                      & 0.003                                                                                   & 0.008                                                                          & 0.000                                                                            & 0.000                                                                          \\ \cline{1-1} \cline{3-10} 
3.00                                                                                 &                                 & 0.02                             & 21.870                                                                                       & 0.351                                                                                         & 0.035                                                                                      & 0.003                                                                                   & 0.010                                                                          & 0.000                                                                            & 0.000                                                                          \\ \hline
0.05                                                                                 & \multirow{15}{*}{1.5}           & 0.96                             & 0.281                                                                                        & 0.965                                                                                         & 0.952                                                                                      & 0.977                                                                                   & 0.990                                                                          & 0.989                                                                            & 0.953                                                                          \\ \cline{1-1} \cline{3-10} 
0.10                                                                                 &                                 & 0.94                             & 0.373                                                                                        & 0.953                                                                                         & 0.938                                                                                      & 0.948                                                                                   & 0.973                                                                          & 0.987                                                                            & 0.885                                                                          \\ \cline{1-1} \cline{3-10} 
0.20                                                                                 &                                 & 0.85                             & 0.587                                                                                        & 0.882                                                                                         & 0.823                                                                                      & 0.845                                                                                   & 0.935                                                                          & 0.991                                                                            & 0.610                                                                          \\ \cline{1-1} \cline{3-10} 
0.30                                                                                 &                                 & 0.67                             & 1.029                                                                                        & 0.684                                                                                         & 0.659                                                                                      & 0.640                                                                                   & 0.842                                                                          & 0.940                                                                            & 0.139                                                                          \\ \cline{1-1} \cline{3-10} 
0.40                                                                                 &                                 & 0.36                             & 1.643                                                                                        & 0.428                                                                                         & 0.345                                                                                      & 0.348                                                                                   & 0.609                                                                          & 0.405                                                                            & 0.030                                                                          \\ \cline{1-1} \cline{3-10} 
0.50                                                                                 &                                 & 0.23                             & 2.801                                                                                        & 0.296                                                                                         & 0.204                                                                                      & 0.192                                                                                   & 0.415                                                                          & 0.152                                                                            & 0.009                                                                          \\ \cline{1-1} \cline{3-10} 
0.60                                                                                 &                                 & 0.13                             & 9.459                                                                                        & 0.155                                                                                         & 0.172                                                                                      & 0.090                                                                                   & 0.225                                                                          & 0.027                                                                            & 0.018                                                                          \\ \cline{1-1} \cline{3-10} 
0.70                                                                                 &                                 & 0.11                             & 6.404                                                                                        & 0.243                                                                                         & 0.119                                                                                      & 0.089                                                                                   & 0.243                                                                          & 0.020                                                                            & 0.004                                                                          \\ \cline{1-1} \cline{3-10} 
0.80                                                                                 &                                 & 0.07                             & 7.845                                                                                        & 0.727                                                                                         & 0.079                                                                                      & 0.038                                                                                   & 0.106                                                                          & 0.005                                                                            & 0.003                                                                          \\ \cline{1-1} \cline{3-10} 
0.90                                                                                 &                                 & 0.06                             & 8.336                                                                                        & 0.718                                                                                         & 0.069                                                                                      & 0.030                                                                                   & 0.077                                                                          & 0.003                                                                            & 0.008                                                                          \\ \cline{1-1} \cline{3-10} 
1.00                                                                                 &                                 & 0.04                             & 9.506                                                                                        & 0.703                                                                                         & 0.052                                                                                      & 0.016                                                                                   & 0.047                                                                          & 0.001                                                                            & 0.000                                                                          \\ \cline{1-1} \cline{3-10} 
1.50                                                                                 &                                 & 0.03                             & 12.377                                                                                       & 0.690                                                                                         & 0.047                                                                                      & 0.008                                                                                   & 0.023                                                                          & 0.000                                                                            & 0.000                                                                          \\ \cline{1-1} \cline{3-10} 
2.00                                                                                 &                                 & 0.02                             & 15.520                                                                                       & 0.685                                                                                         & 0.032                                                                                      & 0.005                                                                                   & 0.015                                                                          & 0.000                                                                            & 0.000                                                                          \\ \cline{1-1} \cline{3-10} 
2.50                                                                                 &                                 & 0.01                             & 20.266                                                                                       & 0.676                                                                                         & 0.031                                                                                      & 0.004                                                                                   & 0.011                                                                          & 0.000                                                                            & 0.000                                                                          \\ \cline{1-1} \cline{3-10} 
3.00                                                                                 &                                 & 0.02                             & 27.568                                                                                       & 0.345                                                                                         & 0.035                                                                                      & 0.002                                                                                   & 0.006                                                                          & 0.000                                                                            & 0.000                                                                          \\ \hline
\end{tabular}}
\caption{Object localisation and counting results for circular aperture, 1280 resolution input images} 
\end{table}

%% file: table_cass.tex
\begin{table}[]
\resizebox{\textwidth}{!}{
\begin{tabular}{|l|l|l|l|l|l|l|l|l|l|}
\hline
\multicolumn{1}{|c|}{\textbf{\begin{tabular}[c]{@{}c@{}}GSD \\ (m/px)\end{tabular}}} & \multicolumn{1}{c|}{\textbf{Q}} & \multicolumn{1}{c|}{\textbf{F1}} & \multicolumn{1}{c|}{\textbf{\begin{tabular}[c]{@{}c@{}}Mean Abs\\ Count Error\end{tabular}}} & \multicolumn{1}{c|}{\textbf{\begin{tabular}[c]{@{}c@{}}Precision\\ All Classes\end{tabular}}} & \multicolumn{1}{c|}{\textbf{\begin{tabular}[c]{@{}c@{}}Recall\\ All Classes\end{tabular}}} & \multicolumn{1}{c|}{\textbf{\begin{tabular}[c]{@{}c@{}}mAP\\ All Classes\end{tabular}}} & \multicolumn{1}{c|}{\textbf{\begin{tabular}[c]{@{}c@{}}AP\\ Cow\end{tabular}}} & \multicolumn{1}{c|}{\textbf{\begin{tabular}[c]{@{}c@{}}AP\\ Sheep\end{tabular}}} & \multicolumn{1}{c|}{\textbf{\begin{tabular}[c]{@{}c@{}}AP\\ Dog\end{tabular}}} \\ \hline
0.05                                                                                 & \multirow{15}{*}{0.5}           & 0.96                             & 0.327                                                                                        & 0.955                                                                                         & 0.956                                                                                      & 0.963                                                                                   & 0.980                                                                          & 0.978                                                                            & 0.931                                                                          \\ \cline{1-1} \cline{3-10} 
0.10                                                                                 &                                 & 0.92                             & 0.453                                                                                        & 0.914                                                                                         & 0.929                                                                                      & 0.931                                                                                   & 0.956                                                                          & 0.982                                                                            & 0.856                                                                          \\ \cline{1-1} \cline{3-10} 
0.20                                                                                 &                                 & 0.82                             & 1.044                                                                                        & 0.866                                                                                         & 0.776                                                                                      & 0.811                                                                                   & 0.887                                                                          & 0.979                                                                            & 0.566                                                                          \\ \cline{1-1} \cline{3-10} 
0.30                                                                                 &                                 & 0.66                             & 1.271                                                                                        & 0.775                                                                                         & 0.604                                                                                      & 0.641                                                                                   & 0.813                                                                          & 0.894                                                                            & 0.216                                                                          \\ \cline{1-1} \cline{3-10} 
0.40                                                                                 &                                 & 0.34                             & 2.939                                                                                        & 0.345                                                                                         & 0.351                                                                                      & 0.290                                                                                   & 0.458                                                                          & 0.372                                                                            & 0.039                                                                          \\ \cline{1-1} \cline{3-10} 
0.50                                                                                 &                                 & 0.21                             & 5.228                                                                                        & 0.259                                                                                         & 0.210                                                                                      & 0.163                                                                                   & 0.332                                                                          & 0.147                                                                            & 0.011                                                                          \\ \cline{1-1} \cline{3-10} 
0.60                                                                                 &                                 & 0.13                             & 12.219                                                                                       & 0.155                                                                                         & 0.177                                                                                      & 0.087                                                                                   & 0.210                                                                          & 0.044                                                                            & 0.006                                                                          \\ \cline{1-1} \cline{3-10} 
0.70                                                                                 &                                 & 0.1                              & 31.082                                                                                       & 0.075                                                                                         & 0.220                                                                                      & 0.068                                                                                   & 0.184                                                                          & 0.016                                                                            & 0.003                                                                          \\ \cline{1-1} \cline{3-10} 
0.80                                                                                 &                                 & 0.06                             & 8.787                                                                                        & 0.714                                                                                         & 0.069                                                                                      & 0.031                                                                                   & 0.081                                                                          & 0.009                                                                            & 0.004                                                                          \\ \cline{1-1} \cline{3-10} 
0.90                                                                                 &                                 & 0.05                             & 8.474                                                                                        & 0.713                                                                                         & 0.065                                                                                      & 0.025                                                                                   & 0.070                                                                          & 0.002                                                                            & 0.003                                                                          \\ \cline{1-1} \cline{3-10} 
1.00                                                                                 &                                 & 0.04                             & 9.088                                                                                        & 0.709                                                                                         & 0.040                                                                                      & 0.015                                                                                   & 0.043                                                                          & 0.002                                                                            & 0.000                                                                          \\ \cline{1-1} \cline{3-10} 
1.50                                                                                 &                                 & 0.03                             & 10.825                                                                                       & 0.692                                                                                         & 0.026                                                                                      & 0.005                                                                                   & 0.014                                                                          & 0.000                                                                            & 0.000                                                                          \\ \cline{1-1} \cline{3-10} 
2.00                                                                                 &                                 & 0.02                             & 13.222                                                                                       & 0.689                                                                                         & 0.027                                                                                      & 0.005                                                                                   & 0.014                                                                          & 0.000                                                                            & 0.000                                                                          \\ \cline{1-1} \cline{3-10} 
2.50                                                                                 &                                 & 0.02                             & 14.370                                                                                       & 0.353                                                                                         & 0.026                                                                                      & 0.004                                                                                   & 0.011                                                                          & 0.000                                                                            & 0.000                                                                          \\ \cline{1-1} \cline{3-10} 
3.00                                                                                 &                                 & 0                                & 14.384                                                                                       & 0.337                                                                                         & 0.005                                                                                      & 0.000                                                                                   & 0.000                                                                          & 0.000                                                                            & 0.000                                                                          \\ \hline
0.05                                                                                 & \multirow{15}{*}{1.0}           & 0.93                             & 0.377                                                                                        & 0.948                                                                                         & 0.921                                                                                      & 0.933                                                                                   & 0.979                                                                          & 0.985                                                                            & 0.835                                                                          \\ \cline{1-1} \cline{3-10} 
0.10                                                                                 &                                 & 0.93                             & 0.460                                                                                        & 0.946                                                                                         & 0.909                                                                                      & 0.933                                                                                   & 0.960                                                                          & 0.983                                                                            & 0.856                                                                          \\ \cline{1-1} \cline{3-10} 
0.20                                                                                 &                                 & 0.78                             & 0.749                                                                                        & 0.798                                                                                         & 0.773                                                                                      & 0.764                                                                                   & 0.898                                                                          & 0.975                                                                            & 0.420                                                                          \\ \cline{1-1} \cline{3-10} 
0.30                                                                                 &                                 & 0.64                             & 1.228                                                                                        & 0.717                                                                                         & 0.595                                                                                      & 0.595                                                                                   & 0.744                                                                          & 0.839                                                                            & 0.204                                                                          \\ \cline{1-1} \cline{3-10} 
0.40                                                                                 &                                 & 0.27                             & 3.050                                                                                        & 0.367                                                                                         & 0.234                                                                                      & 0.219                                                                                   & 0.399                                                                          & 0.237                                                                            & 0.020                                                                          \\ \cline{1-1} \cline{3-10} 
0.50                                                                                 &                                 & 0.16                             & 10.284                                                                                       & 0.162                                                                                         & 0.213                                                                                      & 0.115                                                                                   & 0.256                                                                          & 0.083                                                                            & 0.007                                                                          \\ \cline{1-1} \cline{3-10} 
0.60                                                                                 &                                 & 0.09                             & 22.515                                                                                       & 0.081                                                                                         & 0.178                                                                                      & 0.058                                                                                   & 0.156                                                                          & 0.014                                                                            & 0.003                                                                          \\ \cline{1-1} \cline{3-10} 
0.70                                                                                 &                                 & 0.08                             & 6.266                                                                                        & 0.413                                                                                         & 0.072                                                                                      & 0.054                                                                                   & 0.143                                                                          & 0.013                                                                            & 0.006                                                                          \\ \cline{1-1} \cline{3-10} 
0.80                                                                                 &                                 & 0.05                             & 8.161                                                                                        & 0.711                                                                                         & 0.052                                                                                      & 0.023                                                                                   & 0.063                                                                          & 0.004                                                                            & 0.000                                                                          \\ \cline{1-1} \cline{3-10} 
0.90                                                                                 &                                 & 0.04                             & 12.070                                                                                       & 0.697                                                                                         & 0.062                                                                                      & 0.016                                                                                   & 0.044                                                                          & 0.003                                                                            & 0.000                                                                          \\ \cline{1-1} \cline{3-10} 
1.00                                                                                 &                                 & 0.04                             & 8.269                                                                                        & 0.704                                                                                         & 0.039                                                                                      & 0.013                                                                                   & 0.038                                                                          & 0.001                                                                            & 0.000                                                                          \\ \cline{1-1} \cline{3-10} 
1.50                                                                                 &                                 & 0.02                             & 11.193                                                                                       & 0.683                                                                                         & 0.023                                                                                      & 0.003                                                                                   & 0.010                                                                          & 0.000                                                                            & 0.000                                                                          \\ \cline{1-1} \cline{3-10} 
2.00                                                                                 &                                 & 0.02                             & 11.588                                                                                       & 0.356                                                                                         & 0.022                                                                                      & 0.005                                                                                   & 0.015                                                                          & 0.000                                                                            & 0.000                                                                          \\ \cline{1-1} \cline{3-10} 
2.50                                                                                 &                                 & 0.02                             & 14.398                                                                                       & 0.351                                                                                         & 0.025                                                                                      & 0.003                                                                                   & 0.008                                                                          & 0.000                                                                            & 0.000                                                                          \\ \cline{1-1} \cline{3-10} 
3.00                                                                                 &                                 & 0.01                             & 14.061                                                                                       & 0.346                                                                                         & 0.016                                                                                      & 0.003                                                                                   & 0.008                                                                          & 0.000                                                                            & 0.000                                                                          \\ \hline
0.05                                                                                 & \multirow{15}{*}{1.5}           & 0.93                             & 0.425                                                                                        & 0.922                                                                                         & 0.931                                                                                      & 0.942                                                                                   & 0.972                                                                          & 0.986                                                                            & 0.867                                                                          \\ \cline{1-1} \cline{3-10} 
0.10                                                                                 &                                 & 0.91                             & 0.479                                                                                        & 0.934                                                                                         & 0.896                                                                                      & 0.920                                                                                   & 0.956                                                                          & 0.987                                                                            & 0.819                                                                          \\ \cline{1-1} \cline{3-10} 
0.20                                                                                 &                                 & 0.77                             & 0.859                                                                                        & 0.801                                                                                         & 0.741                                                                                      & 0.767                                                                                   & 0.876                                                                          & 0.975                                                                            & 0.451                                                                          \\ \cline{1-1} \cline{3-10} 
0.30                                                                                 &                                 & 0.52                             & 1.778                                                                                        & 0.537                                                                                         & 0.527                                                                                      & 0.509                                                                                   & 0.674                                                                          & 0.750                                                                            & 0.102                                                                          \\ \cline{1-1} \cline{3-10} 
0.40                                                                                 &                                 & 0.24                             & 4.137                                                                                        & 0.298                                                                                         & 0.223                                                                                      & 0.201                                                                                   & 0.404                                                                          & 0.189                                                                            & 0.010                                                                          \\ \cline{1-1} \cline{3-10} 
0.50                                                                                 &                                 & 0.15                             & 8.924                                                                                        & 0.152                                                                                         & 0.188                                                                                      & 0.104                                                                                   & 0.253                                                                          & 0.055                                                                            & 0.003                                                                          \\ \cline{1-1} \cline{3-10} 
0.60                                                                                 &                                 & 0.07                             & 7.386                                                                                        & 0.397                                                                                         & 0.076                                                                                      & 0.040                                                                                   & 0.108                                                                          & 0.010                                                                            & 0.002                                                                          \\ \cline{1-1} \cline{3-10} 
0.70                                                                                 &                                 & 0.07                             & 6.877                                                                                        & 0.733                                                                                         & 0.074                                                                                      & 0.043                                                                                   & 0.119                                                                          & 0.009                                                                            & 0.001                                                                          \\ \cline{1-1} \cline{3-10} 
0.80                                                                                 &                                 & 0.05                             & 9.708                                                                                        & 0.704                                                                                         & 0.058                                                                                      & 0.018                                                                                   & 0.053                                                                          & 0.001                                                                            & 0.000                                                                          \\ \cline{1-1} \cline{3-10} 
0.90                                                                                 &                                 & 0.04                             & 8.114                                                                                        & 0.702                                                                                         & 0.043                                                                                      & 0.013                                                                                   & 0.036                                                                          & 0.002                                                                            & 0.000                                                                          \\ \cline{1-1} \cline{3-10} 
1.00                                                                                 &                                 & 0.04                             & 8.213                                                                                        & 0.701                                                                                         & 0.037                                                                                      & 0.011                                                                                   & 0.032                                                                          & 0.000                                                                            & 0.000                                                                          \\ \cline{1-1} \cline{3-10} 
1.50                                                                                 &                                 & 0.02                             & 14.567                                                                                       & 0.349                                                                                         & 0.025                                                                                      & 0.004                                                                                   & 0.012                                                                          & 0.000                                                                            & 0.000                                                                          \\ \cline{1-1} \cline{3-10} 
2.00                                                                                 &                                 & 0.02                             & 18.939                                                                                       & 0.011                                                                                         & 0.026                                                                                      & 0.002                                                                                   & 0.007                                                                          & 0.000                                                                            & 0.000                                                                          \\ \cline{1-1} \cline{3-10} 
2.50                                                                                 &                                 & 0.02                             & 14.339                                                                                       & 0.682                                                                                         & 0.021                                                                                      & 0.003                                                                                   & 0.008                                                                          & 0.000                                                                            & 0.000                                                                          \\ \cline{1-1} \cline{3-10} 
3.00                                                                                 &                                 & 0.01                             & 14.246                                                                                       & 0.342                                                                                         & 0.012                                                                                      & 0.001                                                                                   & 0.002                                                                          & 0.000                                                                            & 0.000                                                                          \\ \hline
\end{tabular}}
\caption{Object localisation and counting results for cassegrain aperture, 640 resolution input images} 
\end{table}

%% file: table_cassHighres.tex
\begin{table}[]
\resizebox{\textwidth}{!}{
\begin{tabular}{|l|l|l|l|l|l|l|l|l|l|}
\hline
\multicolumn{1}{|c|}{\textbf{\begin{tabular}[c]{@{}c@{}}GSD \\ (m/px)\end{tabular}}} & \multicolumn{1}{c|}{\textbf{Q}} & \multicolumn{1}{c|}{\textbf{F1}} & \multicolumn{1}{c|}{\textbf{\begin{tabular}[c]{@{}c@{}}Mean Abs\\ Count Error\end{tabular}}} & \multicolumn{1}{c|}{\textbf{\begin{tabular}[c]{@{}c@{}}Precision\\ All Classes\end{tabular}}} & \multicolumn{1}{c|}{\textbf{\begin{tabular}[c]{@{}c@{}}Recall\\ All Classes\end{tabular}}} & \multicolumn{1}{c|}{\textbf{\begin{tabular}[c]{@{}c@{}}mAP\\ All Classes\end{tabular}}} & \multicolumn{1}{c|}{\textbf{\begin{tabular}[c]{@{}c@{}}AP\\ Cow\end{tabular}}} & \multicolumn{1}{c|}{\textbf{\begin{tabular}[c]{@{}c@{}}AP\\ Sheep\end{tabular}}} & \multicolumn{1}{c|}{\textbf{\begin{tabular}[c]{@{}c@{}}AP\\ Dog\end{tabular}}} \\ \hline
0.05                                                                                 & \multirow{15}{*}{0.5}           & 0.96                             & 0.278                                                                                        & 0.966                                                                                         & 0.950                                                                                      & 0.973                                                                                   & 0.985                                                                          & 0.988                                                                            & 0.948                                                                          \\ \cline{1-1} \cline{3-10} 
0.10                                                                                 &                                 & 0.94                             & 0.397                                                                                        & 0.966                                                                                         & 0.919                                                                                      & 0.956                                                                                   & 0.975                                                                          & 0.994                                                                            & 0.898                                                                          \\ \cline{1-1} \cline{3-10} 
0.20                                                                                 &                                 & 0.84                             & 0.541                                                                                        & 0.870                                                                                         & 0.812                                                                                      & 0.821                                                                                   & 0.928                                                                          & 0.992                                                                            & 0.544                                                                          \\ \cline{1-1} \cline{3-10} 
0.30                                                                                 &                                 & 0.68                             & 1.278                                                                                        & 0.796                                                                                         & 0.613                                                                                      & 0.666                                                                                   & 0.830                                                                          & 0.934                                                                            & 0.235                                                                          \\ \cline{1-1} \cline{3-10} 
0.40                                                                                 &                                 & 0.36                             & 1.871                                                                                        & 0.356                                                                                         & 0.372                                                                                      & 0.317                                                                                   & 0.564                                                                          & 0.367                                                                            & 0.020                                                                          \\ \cline{1-1} \cline{3-10} 
0.50                                                                                 &                                 & 0.24                             & 5.105                                                                                        & 0.238                                                                                         & 0.255                                                                                      & 0.193                                                                                   & 0.407                                                                          & 0.164                                                                            & 0.009                                                                          \\ \cline{1-1} \cline{3-10} 
0.60                                                                                 &                                 & 0.15                             & 10.061                                                                                       & 0.144                                                                                         & 0.192                                                                                      & 0.098                                                                                   & 0.244                                                                          & 0.041                                                                            & 0.009                                                                          \\ \cline{1-1} \cline{3-10} 
0.70                                                                                 &                                 & 0.12                             & 7.699                                                                                        & 0.236                                                                                         & 0.139                                                                                      & 0.092                                                                                   & 0.251                                                                          & 0.013                                                                            & 0.010                                                                          \\ \cline{1-1} \cline{3-10} 
0.80                                                                                 &                                 & 0.1                              & 27.260                                                                                       & 0.126                                                                                         & 0.170                                                                                      & 0.048                                                                                   & 0.107                                                                          & 0.005                                                                            & 0.033                                                                          \\ \cline{1-1} \cline{3-10} 
0.90                                                                                 &                                 & 0.06                             & 8.459                                                                                        & 0.721                                                                                         & 0.075                                                                                      & 0.030                                                                                   & 0.085                                                                          & 0.003                                                                            & 0.003                                                                          \\ \cline{1-1} \cline{3-10} 
1.00                                                                                 &                                 & 0.04                             & 10.997                                                                                       & 0.698                                                                                         & 0.055                                                                                      & 0.013                                                                                   & 0.037                                                                          & 0.001                                                                            & 0.000                                                                          \\ \cline{1-1} \cline{3-10} 
1.50                                                                                 &                                 & 0.03                             & 10.956                                                                                       & 0.695                                                                                         & 0.030                                                                                      & 0.007                                                                                   & 0.020                                                                          & 0.000                                                                            & 0.000                                                                          \\ \cline{1-1} \cline{3-10} 
2.00                                                                                 &                                 & 0.03                             & 12.801                                                                                       & 0.690                                                                                         & 0.029                                                                                      & 0.007                                                                                   & 0.021                                                                          & 0.000                                                                            & 0.000                                                                          \\ \cline{1-1} \cline{3-10} 
2.50                                                                                 &                                 & 0.01                             & 19.605                                                                                       & 0.342                                                                                         & 0.021                                                                                      & 0.001                                                                                   & 0.003                                                                          & 0.000                                                                            & 0.000                                                                          \\ \cline{1-1} \cline{3-10} 
3.00                                                                                 &                                 & 0.02                             & 15.132                                                                                       & 0.351                                                                                         & 0.030                                                                                      & 0.005                                                                                   & 0.014                                                                          & 0.000                                                                            & 0.000                                                                          \\ \hline
0.05                                                                                 & \multirow{15}{*}{1.0}           & 0.97                             & 0.353                                                                                        & 0.974                                                                                         & 0.967                                                                                      & 0.984                                                                                   & 0.990                                                                          & 0.988                                                                            & 0.975                                                                          \\ \cline{1-1} \cline{3-10} 
0.10                                                                                 &                                 & 0.94                             & 0.424                                                                                        & 0.946                                                                                         & 0.931                                                                                      & 0.952                                                                                   & 0.974                                                                          & 0.991                                                                            & 0.891                                                                          \\ \cline{1-1} \cline{3-10} 
0.20                                                                                 &                                 & 0.82                             & 0.684                                                                                        & 0.866                                                                                         & 0.781                                                                                      & 0.801                                                                                   & 0.911                                                                          & 0.988                                                                            & 0.503                                                                          \\ \cline{1-1} \cline{3-10} 
0.30                                                                                 &                                 & 0.62                             & 1.070                                                                                        & 0.738                                                                                         & 0.583                                                                                      & 0.616                                                                                   & 0.804                                                                          & 0.873                                                                            & 0.170                                                                          \\ \cline{1-1} \cline{3-10} 
0.40                                                                                 &                                 & 0.3                              & 2.348                                                                                        & 0.355                                                                                         & 0.276                                                                                      & 0.266                                                                                   & 0.506                                                                          & 0.269                                                                            & 0.022                                                                          \\ \cline{1-1} \cline{3-10} 
0.50                                                                                 &                                 & 0.2                              & 4.108                                                                                        & 0.245                                                                                         & 0.190                                                                                      & 0.141                                                                                   & 0.329                                                                          & 0.073                                                                            & 0.020                                                                          \\ \cline{1-1} \cline{3-10} 
0.60                                                                                 &                                 & 0.11                             & 4.994                                                                                        & 0.501                                                                                         & 0.099                                                                                      & 0.074                                                                                   & 0.194                                                                          & 0.019                                                                            & 0.011                                                                          \\ \cline{1-1} \cline{3-10} 
0.70                                                                                 &                                 & 0.09                             & 6.000                                                                                        & 0.795                                                                                         & 0.068                                                                                      & 0.071                                                                                   & 0.194                                                                          & 0.013                                                                            & 0.005                                                                          \\ \cline{1-1} \cline{3-10} 
0.80                                                                                 &                                 & 0.06                             & 10.605                                                                                       & 0.713                                                                                         & 0.084                                                                                      & 0.035                                                                                   & 0.084                                                                          & 0.003                                                                            & 0.017                                                                          \\ \cline{1-1} \cline{3-10} 
0.90                                                                                 &                                 & 0.05                             & 8.886                                                                                        & 0.707                                                                                         & 0.053                                                                                      & 0.019                                                                                   & 0.054                                                                          & 0.002                                                                            & 0.001                                                                          \\ \cline{1-1} \cline{3-10} 
1.00                                                                                 &                                 & 0.04                             & 8.064                                                                                        & 0.705                                                                                         & 0.034                                                                                      & 0.012                                                                                   & 0.036                                                                          & 0.000                                                                            & 0.000                                                                          \\ \cline{1-1} \cline{3-10} 
1.50                                                                                 &                                 & 0.03                             & 11.494                                                                                       & 0.694                                                                                         & 0.029                                                                                      & 0.008                                                                                   & 0.023                                                                          & 0.000                                                                            & 0.000                                                                          \\ \cline{1-1} \cline{3-10} 
2.00                                                                                 &                                 & 0.02                             & 22.706                                                                                       & 0.345                                                                                         & 0.026                                                                                      & 0.002                                                                                   & 0.006                                                                          & 0.000                                                                            & 0.000                                                                          \\ \cline{1-1} \cline{3-10} 
2.50                                                                                 &                                 & 0.02                             & 15.094                                                                                       & 0.347                                                                                         & 0.025                                                                                      & 0.003                                                                                   & 0.009                                                                          & 0.000                                                                            & 0.000                                                                          \\ \cline{1-1} \cline{3-10} 
3.00                                                                                 &                                 & 0.01                             & 15.161                                                                                       & 0.343                                                                                         & 0.017                                                                                      & 0.001                                                                                   & 0.004                                                                          & 0.000                                                                            & 0.000                                                                          \\ \hline
0.05                                                                                 & \multirow{15}{*}{1.5}           & 0.95                             & 0.301                                                                                        & 0.968                                                                                         & 0.942                                                                                      & 0.974                                                                                   & 0.984                                                                          & 0.989                                                                            & 0.949                                                                          \\ \cline{1-1} \cline{3-10} 
0.10                                                                                 &                                 & 0.93                             & 0.476                                                                                        & 0.962                                                                                         & 0.905                                                                                      & 0.939                                                                                   & 0.970                                                                          & 0.991                                                                            & 0.857                                                                          \\ \cline{1-1} \cline{3-10} 
0.20                                                                                 &                                 & 0.81                             & 0.611                                                                                        & 0.847                                                                                         & 0.787                                                                                      & 0.790                                                                                   & 0.920                                                                          & 0.982                                                                            & 0.468                                                                          \\ \cline{1-1} \cline{3-10} 
0.30                                                                                 &                                 & 0.54                             & 1.681                                                                                        & 0.554                                                                                         & 0.533                                                                                      & 0.539                                                                                   & 0.744                                                                          & 0.787                                                                            & 0.085                                                                          \\ \cline{1-1} \cline{3-10} 
0.40                                                                                 &                                 & 0.28                             & 3.295                                                                                        & 0.356                                                                                         & 0.254                                                                                      & 0.236                                                                                   & 0.453                                                                          & 0.232                                                                            & 0.022                                                                          \\ \cline{1-1} \cline{3-10} 
0.50                                                                                 &                                 & 0.14                             & 4.705                                                                                        & 0.404                                                                                         & 0.122                                                                                      & 0.113                                                                                   & 0.289                                                                          & 0.038                                                                            & 0.013                                                                          \\ \cline{1-1} \cline{3-10} 
0.60                                                                                 &                                 & 0.08                             & 8.173                                                                                        & 0.739                                                                                         & 0.103                                                                                      & 0.061                                                                                   & 0.167                                                                          & 0.010                                                                            & 0.006                                                                          \\ \cline{1-1} \cline{3-10} 
0.70                                                                                 &                                 & 0.07                             & 6.553                                                                                        & 0.740                                                                                         & 0.075                                                                                      & 0.051                                                                                   & 0.133                                                                          & 0.004                                                                            & 0.015                                                                          \\ \cline{1-1} \cline{3-10} 
0.80                                                                                 &                                 & 0.05                             & 8.453                                                                                        & 0.712                                                                                         & 0.061                                                                                      & 0.024                                                                                   & 0.067                                                                          & 0.003                                                                            & 0.000                                                                          \\ \cline{1-1} \cline{3-10} 
0.90                                                                                 &                                 & 0.05                             & 8.871                                                                                        & 0.706                                                                                         & 0.057                                                                                      & 0.018                                                                                   & 0.051                                                                          & 0.001                                                                            & 0.001                                                                          \\ \cline{1-1} \cline{3-10} 
1.00                                                                                 &                                 & 0.04                             & 9.848                                                                                        & 0.702                                                                                         & 0.042                                                                                      & 0.013                                                                                   & 0.036                                                                          & 0.000                                                                            & 0.001                                                                          \\ \cline{1-1} \cline{3-10} 
1.50                                                                                 &                                 & 0.02                             & 15.958                                                                                       & 0.352                                                                                         & 0.028                                                                                      & 0.004                                                                                   & 0.012                                                                          & 0.000                                                                            & 0.000                                                                          \\ \cline{1-1} \cline{3-10} 
2.00                                                                                 &                                 & 0.02                             & 13.701                                                                                       & 0.348                                                                                         & 0.019                                                                                      & 0.002                                                                                   & 0.007                                                                          & 0.000                                                                            & 0.000                                                                          \\ \cline{1-1} \cline{3-10} 
2.50                                                                                 &                                 & 0.02                             & 13.184                                                                                       & 0.352                                                                                         & 0.016                                                                                      & 0.003                                                                                   & 0.009                                                                          & 0.000                                                                            & 0.000                                                                          \\ \cline{1-1} \cline{3-10} 
3.00                                                                                 &                                 & 0                                & 11.789                                                                                       & 0.337                                                                                         & 0.004                                                                                      & 0.000                                                                                   & 0.001                                                                          & 0.000                                                                            & 0.000                                                                          \\ \hline
\end{tabular}}
\caption{Object localisation and counting results for cassegrain aperture, 1280 resolution input images} 
\end{table}